# Data Analysis and Optimization for Intelligent Transportation in Internet of Things

## Urban Traffic Flow Forecast Based on FastGCRNN

Ya Zhang,[1] Mingming Lu,[1] and Haifeng Li[2]

[1] School of Computer Science and Engineering, Central South University, 410083, Changsha, Hunan, China.

[2] School of Geosciences and Info-Physics, Central South University, 410083, Changsha, Hunan, China.

Correspondence should be addressed to Mingming Lu; mingminglu@csu.edu.cn

## Abstract

Traffic forecasting is an important prerequisite for the application of intelligent transportation system in urban traffic networks. The existing works adopted RNN and CNN/GCN, among which GCRN is the state of art work, to characterize the temporal and spatial correlation of traffic flows. However, it is hard to apply GCRN to the large scale road networks due to high computational complexity. To address this problem, we propose to abstract the road network into a geometric graph and build a Fast Graph Convolution Recurrent Neural Network (FastGCRNN) to model the spatial-temporal dependencies of traffic flow. Specifically, We use FastGCN unit to efficiently capture the topological relationship between the roads and the surrounding roads in the graph with reducing the computational complexity through importance sampling, combine GRU unit to capture the temporal dependency of traffic flow, and embed the spatiotemporal features into Seq2Seq based on the Encoder-Decoder framework. Experiments on large-scale traffic data sets illustrate that the proposed method can greatly reduce computational complexity and memory consumption while maintaining relatively high accuracy.

## Introduction

Traffic forecasting using timely information provided by Internet of Things technology(IoT) is an important prerequisite for the application of intelligent transportation system(ITS)[1] in urban traffic networks, because an accurate and efficient prediction model can be used for travellers to select high-quality reference routes, maximize the utilization of road networks, and provide a basis for the reasonable planning of urban construction departments. However, along with worldwide urbanization, urban road networks have been expanded significantly[2], which brings challenges for traffic forecasting because the corresponding computation complexity will greatly increase due to the expanded road networks[3].

This paper mainly studies the problem of urban traffic forecasting based on the Internet of Things technology(IoT) in large urban road traffic networks. This problem is how to use historical traffic flow data to predict traffic flow data in future timestamps in large urban road traffic networks. In literature, there has been plenty of studies in traffic forecasting, including traffic volume, taxi pick-ups, and traffic in/out flow volume. Initially, numerous statistical based methods, such as Historical Average (HA)[4], Time Series[5], K Nearest Neighbors Algorithm (KNN)[6], and Kalman Filter[7], have been proposed to predict road traffic. However, these models are generally suitable for relatively stable traffic flow, which cannot well reflect the temporal correlation of traffic flow data, nor can they reflect the real-time nature of traffic flow. In order to solve the unstable characteristics of traffic flow data, ARIMA[8] and its variants[9], [10] are used in this field [11]. Although these studies show that the prediction can be improved by considering various other factors, they are still unable to capture the complex nonlinear spatiotemporal correlation. The latest advances in deep learning enable researchers to model complex nonlinear relationships and show promising results in multiple fields. This success has inspired many attempts to use deep learning technology in traffic flow prediction. Recent studies have proposed the use of improved LSTM[12] and GRU[13] to predict traffic flow. Furthermore, considering the influence of spatial structure on the traffic flow of different roads, Li et al. [14], [15] proposed to model the traffic volume of the city as an image and partition the city map (the image) into a large number grid. Within each grid cell, the traffic volume within a period of time can be regarded as a pixel value. Based on that, Li adopted ConvLSTM[16] to model the spatial-temporal correlation among traffic flows, where the convolution operation and the LSTM unit are utilized to model spatial and temporal correlation, respectively. However, the conversion of traffic flow into images loses the spatial topology of urban roads. Li et al. [17] modeled the traffic flow as a diffusion process on a directed graph and captured the spatial dependency using bidirectional random walks on the graph, and the temporal dependency using the encoder-decoder architecture with scheduled sampling. Seo et al.[18] used GCN[19]–[21] to extract the spatial topology of the traffic network and RNN to find dynamic patterns to optimize traffic forecasting. However, GCN suffers from the scalability issue, because it requires a lot of space to maintain the entire graph and embed each node in memory[22]–[26], and it has a very high computational complexity[27].

In order to solve the above problem, we propose to form the road network into a geometric graph, and construct a spatiotemporal graph convolution network based on the abstract graph to capture the spatiotemporal features of traffic flow for prediction. We propose to use GCN as the spatial topology extractor of the model and apply the sampling method[28]–[30] to GCN. The method can put the nodes in the graph into the model in batches, and sample the neighbors of the nodes in each batch, extract the nodes that have a greater impact on the nodes in this batch, and perform convolution operations, which greatly reduces the calculation complexity and memory. Consume, can deal with the big graph effectively, the memory overflow problem is not easy to occur. Then we further combine GRU to extract temporal features to achieve the extraction of spatiotemporal features of traffic flow. Finally, we embed spatiotemporal features into Seq2Seq[31] based on Encoder-Decoder framework for prediction.

## Problem Analysis

Urban traffic flow prediction is based on historical traffic flow sequences, which are highly time-varying, nonlinear, and uncertain. The traffic flow in the road network usually has the following temporal characteristics[32]:

a) Periodicity. Traffic flows change periodically. The time series of traffic flow usually presents a wavy or oscillatory fluctuation around the long-term trend;

b) Trend and trend variability[33]. The time series of traffic flow shows a regular change trend. It will not change randomly, but it will continuously change with time. For example, from spring to summer, the traffic volume of the morning peak will gradually advance. Present a trending change;

c) Continuity. Traffic flow has continuity in time, that is, there is a correlation between the value of traffic flow at different times, especially in adjacent time periods.

At a certain time, traffic flow also has some spatial characteristics, such as the impact of traffic flow upstream and downstream of the road on the current road, the rules of speed limit and traffic flow limit of the same level of road, etc.

In view of these two main influence factors, especially considering the large scale of the road network[34]–[39], which requires a lot of time for spatial calculation, this paper proposes the Fast Graph Convolution Recurrent Neural Network (FastGCRNN).

It uses recurrent neural network to capture the long-term temporal dependency of traffic flow, and the graph convolution neural network (GCN) to capture the spatial correlation among roads in different geographical locations. At the same time, importance sampling is applied to GCN to reduce the computational complexity of large road networks.

## Preliminaries

### Notations

Given an undirected graph $G = (V, E, X)$, where $V = \{v_1, v_2, ..., v_n\}$ is a set of nodes with $|V| = n$, $E \subseteq V \times V$ is a set of edges that can be represented as an adjacency matrix $A \in \{0,1\}^{n \times n}$, and $X = [x_1, x_2, ..., x_n]^T \in \mathbb{R}^{n \times d_{in}}$ is a feature matrix with $x_i$ denoting a feature vector of node $v_i \in V$. $d_{in}$ is the length of the historical time series, and each feature in $x_i$ corresponds to the traffic flow at a certain time. Our target is to obtain the traffic information $Y = [y_1, y_2, ..., y_n]^T \in \mathbb{R}^{n \times d_{out}}$ ($d_{out}$ is the length of traffic flow time series to be predicted) of a certain period of time in the future according to the historical traffic information X.

### Graph Convolution Networks

As a semi-supervised model, GCN can learn the hidden representation of each node. The hidden vectors of all nodes in layer $l + 1$ can be represented recursively by the hidden vectors of layer $l$ as follows.

$$H^{(l+1)} = \sigma\left( \tilde{D}^{-\frac{1}{2}} \tilde{A} \tilde{D}^{-\frac{1}{2}} H^{(l)} W^{(l)} \right) \tag{1}$$

where $\tilde{A} = A + I_n$, $W^{(l)}$ denotes the learnable weight matrix at layer $l$, $\tilde{D}_i = \sum_j \tilde{A}_{ij}$, and $\sigma(\cdot)$ is an activation function, such as ReLu. Initially, $H^{(0)} = X$.

## Fast Graph Convolution Recurrent Neural Network

The traffic flow of a road is affected by the traffic flow of the surrounding roads and the historical traffic flow of the road itself, so the prediction model should consider these two factors. To model the temporal dependency of historical traffic on the road, GRU unit is embedded in the Seq2Seq model based on Encoder-Decoder framework to complete sequence prediction, and to model the spatial correlation among neighbor roads, FastGCN is used in the traffic map of the road network to reduce the computational complexity and improve the efficiency. We integrate a model for quickly extracting spatiotemporal features, so we propose the FastGCRNN (Fast Graph Revolution Recurrent Neural Network) model. The overall architecture of the model is shown in Figure 1.

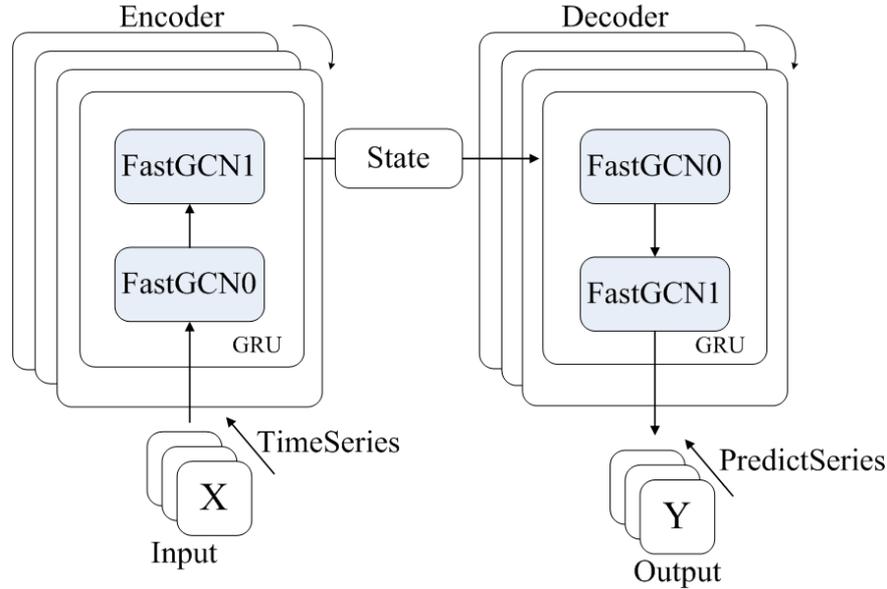

Figure 1: FastGCRNN model.

This model mainly includes six parts, namely:

a) Input sequence X. It is the input data of the whole prediction model, which is fed into the encoder part. In the road network traffic graph, it is the traffic flow of each node in a continuous period of time;

b) Output sequence Y. It is the output of the whole prediction model (the output of decoder part). In the road network traffic graph, it is the traffic flow of each node road in the future;

c) FastGCN unit. It can extract the spatial structure information of the road network through graph convolution. Based on that, it further uses sampling to reduce computational complexity.

d) GRU unit. Traffic flows are time series signals, so we use GRU units to capture the long-term or short-term temporal dependence between the input traffic flow time series, and embeds two FastGCN units in its internal;

e) Encoder unit. It is composed of GRU unit, and the output state of hidden layer is obtained by encoding the time series of the input traffic flow network graph;

f) Decoder unit. It is also composed of GRU units. When it receives the encoder output, the decoder will continuously predict the traffic flow of each node.

The whole FastGCRNN model adopts the Seq2Seq model based on the Encoder-Decoder framework, which can use traffic flow of each road within the road network to predict the future traffic flow. Firstly, the continuous traffic flow data $X$ on the road network is fed into the encoder part, and the data instance at each timestamp needs to go through FastGCN units to extract the spatial structure information between the road nodes, and then it needs to be processed by the GRU units in the encoder to get the temporal features of the traffic flow. After encoding, the hidden state output by the GRU units in the encoder is fed to the GRU units in decoder, and spatial features are further extracted by FastGCN. The final GRU units continuously predicts the traffic flow $Y$.

**Fast spatial feature extractor——FastGCN**

Each road in the urban road network does not exist in isolation, but connects with the surrounding roads to form a whole. The traffic flow between roads is interactive, especially on the two-way road, there are vehicles flowing in and out. To model spatial correlation of traffic flows among road networks, we abstract the roads in road networks as nodes and their intersections as edges, as shown in Figure 2, where blue lines and dots represent road and intersections in road networks, respectively. Since we intend to predict traffic flows of the roads, while GCN can only make prediction on nodes, we model roads as nodes and their intersections as edges, as illustrated through the red triangles and yellow lines in Figure 2, respectively.

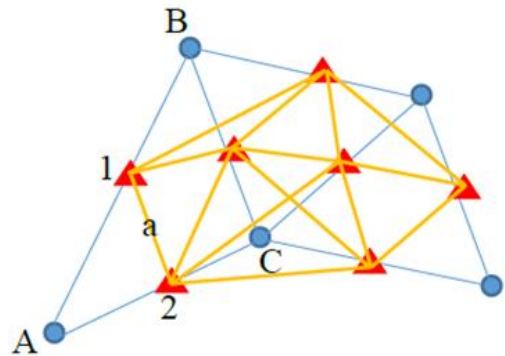

Figure 2: Construction process of road network graph.

In order to consider the influence of multi-hop in GCN, the number of layers of GCN will be increased recursively to realize the information exchange between multiple upstream and downstream roads. However, the recursive neighborhood expansion across layers poses time and memory challenges for training with large, dense graphs. To solve this problem, the FastGCN method is used, which interprets GCN as the integral transformation of the embedded function under the probability measure. The integration at this time can use the Monte Carlo method for consistency estimation, and the node training in the graph can also be performed in batches. Since the node training is carried out in batches, the structure of the graph is not limited, that is, when performing test prediction, the number of nodes and the connection relationship in the graph can change, and it does not have to be the same as the graph structure during training. This increases the generalization ability and scalability of the model to a certain extent.

The nodes in the graph of FastGCN can be regarded as independent and identically distributed sampling points that satisfy a certain probability distribution, and the calculated loss and convolution results are expressed as the integral form of the embedding function of each node. The estimation of integration can be expressed by Monte Carlo approximation which defines the sampling loss and sampling gradient. In order to reduce the variance of estimation, the sampling distribution can be further changed to make it more consistent with the real distribution. For example, the simplest way is to use uniform distribution for sampling convolution. The improved method is to use importance sampling to make it continuously approach the real distribution and reduce the error caused by sampling.

If a node $v$ in the graph G is taken as the observation object, its convolution can be considered as the information embedding expression of node $v$ and all nodes in the graph in the upper layer through the addition of other forms of adjacency matrix, and then the transformation of feature dimension through the trainable parameter matrix, which is equivalent to a discrete integral, and the adjacency matrix is equivalent to a weight given to each node. Therefore, the convolution process of node $v$ in the graph is expressed in integral form as:

$$\tilde{h}^{(l+1)}(v) = \int \tilde{A}(v,u) h^{(l)}(u) W^{(l)} dP(u), h^{(l+1)}(v) = \sigma(\tilde{h}^{(l+1)}(v)), l = 0,...,M-1 \qquad (2)$$

GCN in the form of integration is integrated by Monte Carlo method, and then it is transformed into the discrete form of sampling. At layer $l$, $t_l$ points ($u_1^{(l)}, \cdots, u_{t_l}^{(l)}$) are sampled independently and identically with probability $p$, and the approximate estimation is

$$\tilde{h}_{t_{l+1}}^{(l+1)}(v) := \frac{1}{t_l} \sum_{j=1}^{t_l} \tilde{A}(v,u_j^{(l)}) h_{t_l}^{(l)}(u_j^{(l)}) W^{(l)}, h_{t_{l+1}}^{(l+1)}(v) := \sigma(\tilde{h}_{t_{l+1}}^{(l+1)}(v)), l = 0,...,M-1 \qquad (3)$$

If each layer of convolution uses this method for sampling and information transfer, after layer $M$, the embedded expression of node $v$ is

$$H^{(l+1)}(v,:) = \sigma(\frac{n}{t_l} \sum_{j=1}^{t_l} \tilde{A}(v,u_j^{(l)}) H^{(l)}(u_j^{(l)},:) W^{(l)}), l = 0,...,M-1 \qquad (4)$$

In the above integral form of GCN, the embedded information expression of node V needs to be obtained from all nodes in the graph. However, after sampling, only $t_l$ nodes in the graph need to exchange and fuse information in FastGCN, so the calculation complexity of the whole graph changes from $n^2$ to ($t_l \times n$), and the efficiency is greatly improved.

Here is an example to illustrate the advantages of FastGCN compared with GCN. If the abstract road network graph has 5 nodes and 6 edges, as shown in Figure 3 and Figure 4.

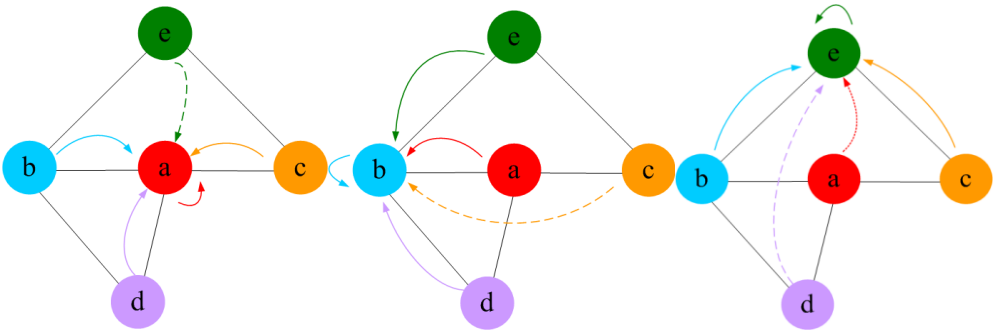

Figure 3: The process of GCN performing a convolution operation.(a) Convolution process of node A.(b) Convolution process of node B (c) Convolution process of node E.

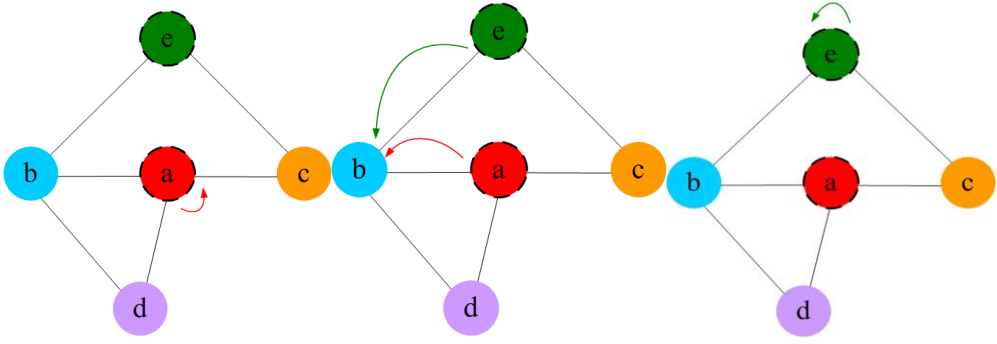

Figure 4:Convolution operation process in a batch of FastGCN under sampling distribution.(a) Sampling convolution operation of node A.(b) Sampling convolution operation of node B. (b) Sampling convolution operation of node E.

In GCN, each epoch must be put into a complete graph, instead of using only a few nodes in the graph, that is, each node in the graph needs to convolute and exchange information with all other nodes in the graph. In FastGCN, we decomposes the large graph into several small graphs by batch operation and puts them into memory, as well as the method of sampling to remove the information exchange with some low correlation nodes. Each node only interacts with the sampled nodes in the graph. As shown in Figure 4, each node only interacts with node A and node E. In this way, the computing efficiency is greatly improved, especially when it can be calculated on a large graph without memory overflow.

For the sampling method, in order to make the sampling closer to the real connected nodes, FastGCN uses not uniform sampling[40], but importance sampling. That is, each node is not sampled according to the same probability, but using probability distribution $Q$. No matter what probability distribution sampling is used, the mean value of the sample is constant, but it will affect the variance of the sample. In order to minimize the error, the distribution $Q$ which can minimize the sample variance is selected here. At this time, the calculation output of node $v$ passing through FastGCN layer is the Formula (5).

$$H^{(l+1)}(v,:) = \sigma(\frac{1}{t_l}\sum_{j=1}^{t_l}\frac{\tilde{A}(v,u_j^{(l)})H^{(l)}(u_j^{(l)},:)W^{(l)}}{q(u_j^{(l)})}), u_j^{(l)} \sim q, l = 0,\ldots,M-1 \qquad (5)$$

In the experiment, only two FastGCN units were used to extract spatial features. This is because we need to avoid the problem of over smoothing[41]. The specific calculation process is as follows:

$$f(\tilde{A},X) = \sigma(\frac{1}{t_l}\sum_{j=1}^{t_l}\frac{\tilde{A}(v,u_j^{(1)})\sigma(\frac{1}{t_l}\sum_{j=1}^{t_l}\frac{\tilde{A}(v,u_j^{(0)})X(u_j^{(0)},:)W^{(0)}}{q(u_j^{(0)})})(u_j^{(1)},:)W^{(1)}}{q(u_j^{(1)})}), u_j^{(0)} \sim q, u_j^{(1)} \sim q \qquad (6)$$

### Fast temporal feature extractor——GRU

This is a key issue to effectively capture the long-term temporal dependence of traffic flow. The observed value of each timestamp is shown in Figure 5. The flow value of each node will change with time. The prediction is a typical time series prediction problem, that is, given the observed value of each road at $d_{in}$ timestamps in history, the traffic flow value of $d_{out}$ timestamps in the future will be predicted.

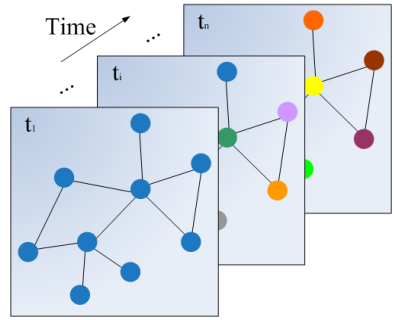

Figure 5:Traffic flow data with graph structure at different timestamp.

LSTM and GRU are commonly used in time series prediction. Both models use gating mechanisms to remember as much long-term information as possible and are equally effective for various tasks. To maximize efficiency, we chose GRU with relatively simple structure, fewer parameters, and faster training ability. GRU unit has update gate, reset gate and memory unit, which can make it have a process of screening memory for historical data, so it can retain long-term memory. In GRU, time sequence information is saved by memory unit, which can capture long and short-term memory in time and improve the accuracy of prediction.

In order to complete the sequence prediction, the Seq2Seq model based on the Encoder-Decoder structure is used. Seq2Seq puts the input history sequence into GRU, extracts the timing features, and obtains the hidden state vector $C$ of the input sequence as the coding result of the encoder. This state vector $C$ contains the feature information of all the previous moments, which is a centralized embodiment of their temporal features. In the decoder, $C$ is used as the initial input of decoder to generate the predicted time series. In this way, Seq2Seq can extract the temporal characteristics of the traffic volume in the previous period, such as the proximity, trend, and periodicity of the traffic flow in the time dimension. When predicting the traffic volume, the model can obtain a smoothly changing traffic volume according to the proximity, and the characteristics of the proximity can be adjusted according to the trend and periodicity.

## Experiment

In order to illustrate the role of the model in the large graph, 1865 roads in Luohu District of Shenzhen city are selected for the experiment, and the specific roads and areas are shown in Figure 6.

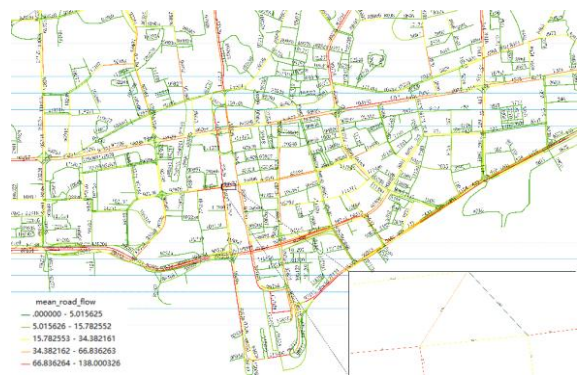

Figure 6:Part of the road network map of Luohu District, Shenzhen.

To calculate the traffic flows in each road, we map the GPS coordinates to the corresponding roads through the Frechet method[42]. The format of the mapped data is shown in Table 1. The core fields are road number(road_id), license plate number(car_id), and upload time(time). Each data record represents the information, the taxis with the car_id is on the road with road_id at the specific time.

Table 1: Shenzhen taxi GPS record information example.

| road_id | car_id | time |
|---------|--------|------|
| 92230 | 02341 | 2015-01-01 00:03:46 |
| 92230 | 03982 | 2015-01-02 06:23:12 |
| … | … | … |

**Data preprocessing**

In data preprocessing, the taxi data in Shenzhen is transformed into the form of continuous time stamps on the road network, i.e. the traffic data shown in Figure 5. Specifically, we map the original GPS upload data to the road, and count the traffic flow on each road in each time period. The data preprocessing algorithm is shown in Algorithm 1.

| **Algorithm 1:** Generate traffic flow time series for different roads |
|---|
| 1:    **Initialize:** *time_interval* = 5min (or 30min), |
|                        *begin_time* = *2015-01-01 00:00:00*, |
|                        *roadflow[roadid][time_num]* = 0 |
| 2:    **For** *Every data record* **do** |
| 3:      *time_num* ←(*time* – *begin_time*) / *time_interval* |
| 4:    **End for** |
| 5:    *All data records* are grouped by *car_id*, sorted by *time_num* within the group |
| 6:    **For** *each group records* **do** |
| 7:      remove duplicates *records* based on *road_id* and *time_num* |
| 8:      count *roadflow* |
| 9:    **End for** |
| 10:   **Output：** *roadflow* |

**Comparative Experiment**

The biggest advantage of FastGCRNN model is that it can be applied to large graphs, and it can reduce the computational complexity without losing the accuracy of the model. On the road network data of Shenzhen, the experiment is conducted with the traffic flow series of different time intervals to compare with some classic traffic flow prediction models (1) HA (2) ARIMA (3) SVR (4) LSTM (5) ConvLSTM (6) GCRN[18] (7) GCRNN-nosample. The evaluation standard used in the experiment is Root Mean Squared Error (RMSE)[43]. The specific experimental results are shown in Table 2.

Table 2: Comparison of results between FastGCRNN model and other traffic flow prediction models.

| Model \ Time | RMSE | |
|---|---|---|
| | 5min | 30min |
| HA | 19.502 | 23.158 |
| ARIMA | 17.541 | 19.097 |
| SVR | 17.895 | 19.005 |
| LSTM | 13.102 | 16.930 |
| ConvLSTM | 19.481 | 21.038 |
| GCRN | 11.892 | 16.265 |
| GCRNN-nosample | 9.950 | **16.231** |
| FastGCRNN | **9.867** | 16.2734 |

From the table results, we can find that FastGCRNN model has reached the best prediction performance in terms of RMSE. In these comparison models, HA, ARMIMA, SVR and LSTM only consider the temporal correlation without considering the spatial correlation, which is also one of the reasons for their poor accuracy. ConvLSTM divides the urban area into a grid, and maps the traffic volume in each time period to the grid, and the traffic volume is regarded as the pixel value of the grid. Although this method considers the spatial correlation of vehicle flow, it also loses the topological structure relationship of the road network graph.

To verify the proposed GCRNN can reduce the computational complexity, compared with the GCRN model, which also captures the topology information of the road network, the result is shown in Figure 7.

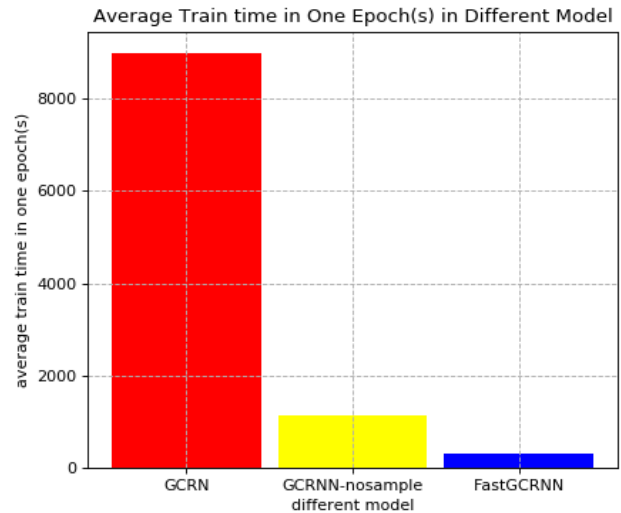

Figure 7: Time consumption of training an epoch with different models.

In Figure 7, we only compare the baselines with higher prediction accuracy, namely, GCRN and GCRNN-nosample. From Figure 7, it can be observed that the computational complexity of FastGCRNN is the lowest. The training time of FastGCRNN is about 0.03 times that of GCRN. Moreover, FastGCRNN reduces the training time to 1/3 times that of GCRNN-nosample, i.e, the GCRNN model without sampling. From the experiment results, it can be concluded that both the GCRNN model and the sampling method can reduced the training time.

**Model parameter analysis**

In FastGCRNN, each sampling point has a certain effect on the accuracy and training time of the model. When using 1685 roads in Shenzhen for experiments, different sampling sizes were set to compare the accuracy and time changes. The experimental results are shown in Figure 8. The abscissa in the figure shows the sampling size of FastGCN unit in the first and second layers respectively. The blue column represents the RMSE of the prediction results. The red line indicates the time consumption in each epoch, and the upper and lower ends are the maximum and minimum values of time consumption in the training process.

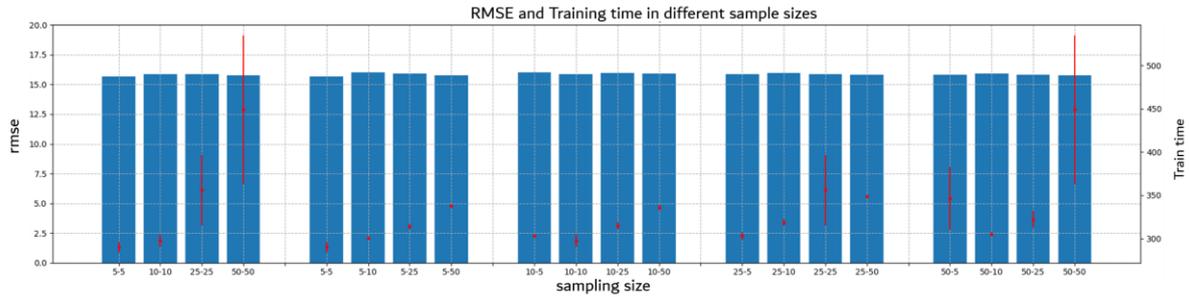

Figure 8: RMSE and training time when using different sampling sizes in two layers of FastGCN

From the experimental results, it can be seen that choosing different sampling sizes has little effect on accuracy, and it does not necessarily mean that the more samples, the more information obtained, the better the prediction effect. For example, the accuracy of sampling 50 nodes for each layer in the figure is not the best, because there is "bridge" type (other nodes affecting the central node will spread to other unrelated distant areas) and "tree" type (other nodes affecting the central node will be limited to the small area to which the node belongs) of connection relationship between nodes[44]. If more nodes are sampled, the influence relationship of the nodes will spread to unrelated areas, resulting in information redundancy, misleading the update of node features, and reducing the prediction accuracy. In addition, in the road network graph, intersections generally connect four roads, that is to say, selecting four nodes in one hop can complete the extraction of feature information. Here is the statistics of 1865 selected roads' degrees, as shown in Figure 9. Among them, the nodes with degree 4 are the most, and the degrees of 70% of the nodes are less than 5, and the degrees of nearly 99% of the nodes are less than 7. Therefore, the case of sampling size 5 can already include the neighbors in one-hop around it. In this case, not only the training time is reduced, but also the accuracy is not reduced.

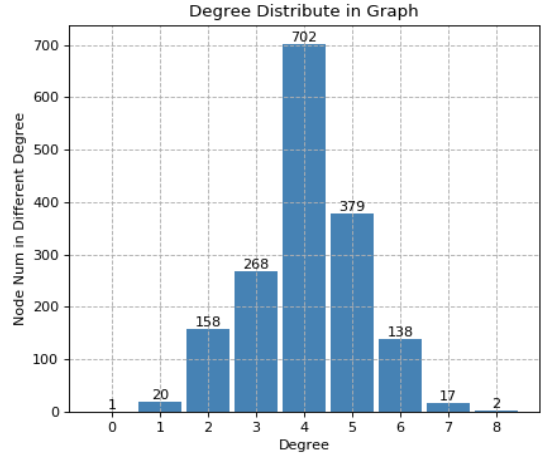

Figure 9: Distribution of node degree of road network graph in Shenzhen.

And we compared the time consumption of FastGCN and standard GCN in different sizes of graphs. The experimental results are shown in Figure 10.

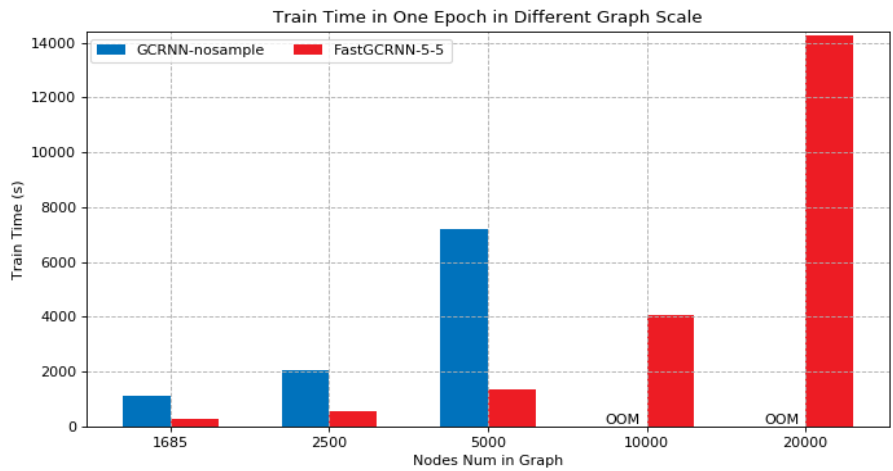

Figure 10: Time consumption of FastGCRNN and GCRNN unsampled models at different graph sizes.

From the experimental results, it can be seen that FastGCRNN has obvious advantages in dealing with large graph problems. Especially when the size of graph reaches a certain degree, FastGCRNN is still running normally when GCRNN-nosample model has overflowed memory and can not be trained.

## Conclusions

This paper mainly deals with the problem of large graphs with spatiotemporal properties by constructing the FastGCRNN model and applies them to road network traffic graphs. The model predicts the traffic flow by extracting the temporal and spatial attributes of the traffic flow on the large-scale road networks. Among them, FastGCN is used to extract the topological structure in the space and accelerate training and reduce complexity. GRU is used to extract time series features, and the Seq2Seq model based on the Encoder-Decoder framework can complete sequence prediction tasks of unequal length. The most prominent advantage of this model is the FastGCN embedded in it, which uses the sampling method to accelerate the extraction of spatial features, reduce computational complexity, and improve efficiency. Moreover, the model is not prone to memory overflow in processing large-scale graph-structured data.

It is worth mentioning that this model is not only applicable to traffic flow data, but also applicable to all graph structure data with spatiotemporal characteristics, especially the larger scale data.

## Data Availability

The data used to support the findings of this study are available upon request to Ya Zhang, zndxxxxyzy@csu.edu.cn.

## Conflicts of Interest

The authors declare that there are no conflicts of interest.